\pdfoutput=1

\documentclass[11pt]{article}

\usepackage[]{emnlp2021}

\usepackage{times}
\usepackage{latexsym}
\usepackage{microtype}
\usepackage{graphicx}
\usepackage{booktabs}  
\usepackage{multirow}
\usepackage{subfigure}
\usepackage{verbatim}
\usepackage{CJKutf8} 
\usepackage{amssymb}
\usepackage{amsmath}

\usepackage{algorithm}
\usepackage{algorithmic}
\usepackage{makecell}
\usepackage[T1]{fontenc}

\usepackage[utf8]{inputenc}

\usepackage{microtype}

\title{NewsBERT: Distilling Pre-trained Language Model for \\ Intelligent News Application}

\author{Chuhan Wu$^1$~~Fangzhao Wu$^2$~~Yang Yu$^3$~~Tao Qi$^1$~~Yongfeng Huang$^1$~~Qi Liu$^3$\\
    $^1$Department of Electronic Engineering \& BNRist, Tsinghua University, Beijing 100084, China  \\
     $^2$Microsoft Research Asia, Beijing 100080, China \\ 
     $^3$University of Science and Technology of China, Hefei 230027, China\\
  {\tt \{wuchuhan15, wufangzhao, taoqi.qt\}@gmail.com} \\
  {\tt tomyu613@icloud.com,} 
  {\tt yfhuang@tsinghua.edu.cn,}
  {\tt qiliuql@ustc.edu.cn}
  }
  
\begin{document}

\maketitle

\begin{abstract}

Pre-trained language models (PLMs) like BERT have made great progress in NLP.
News articles usually contain rich textual information, and PLMs have the potentials to enhance news text modeling for various intelligent news applications like news recommendation and retrieval.
However, most existing PLMs are in huge size with hundreds of millions of parameters.
Many online news applications need to serve millions of users with low latency tolerance, which poses great challenges to incorporating PLMs in these scenarios.
Knowledge distillation techniques can compress a large PLM into a much smaller one and meanwhile keeps good performance.
However, existing language models are pre-trained and distilled on general corpus like Wikipedia, which has gaps with the news domain and may be suboptimal for news intelligence.
In this paper, we propose NewsBERT, which can distill PLMs for efficient and effective news intelligence.
In our approach, we design a teacher-student joint learning and distillation framework to collaboratively learn both teacher and student models, where the student model can learn from the learning experience of the teacher model.
In addition, we propose a momentum distillation method by incorporating the gradients of teacher model into the update of student model to better transfer the knowledge learned by the teacher model.
Thorough experiments on two real-world datasets with three tasks show that NewsBERT can empower various intelligent news applications with much smaller models.

\end{abstract}

\section{Introduction}

Pre-trained language models (PLMs) like BERT~\cite{devlin2019bert} and GPT~\cite{radford2019language} have achieved remarkable success in various NLP applications~\cite{liu2019roberta,yang2019xlnet}.
These PLMs are usually in huge size with hundreds of millions of parameters~\cite{qiu2020pre}.
For example, the BERT-Base model contains about 110M parameters and 12 Transformer~\cite{vaswani2017attention} layers, which may raise a high demand of computational resources in model training and inference.
However, many online applications need to provide  services for a large number of concurrent users and the tolerance of latency is often low, which hinders the deployment of large-scale PLMs in these systems~\cite{sanh2019distilbert}.

In recent years, online news websites such as MSN News and Google News have gained huge popularity for users to digest digital news~\cite{wu2019npa}.
These news websites usually involve a series of intelligent news applications like automatic news topic classification~\cite{wu2019neural}, news headline generation~\cite{tan2017neural} and news recommendation~\cite{okura2017embedding,wang2018dkn,wu2019npa,wu2019,wu2019nrms,wu2021empowering}.
In these applications, text modeling is a critical technique because news articles usually contain rich textual content~\cite{wang2020fine}.
Thus, these applications would benefit a lot from the powerful language understanding ability of PLMs if they could be incorporated in an efficient way, which further has the potential to improve the news reading experience of millions of users~\cite{wu2020mind}.

Knowledge distillation is a technique that can compress a cumbersome teacher model into a lighter-weight student model by transferring useful knowledge~\cite{hinton2015distilling,kim2016sequence}.
It has been employed to compress many huge pre-trained language models into much smaller versions and meanwhile keep most of the original performance~\cite{sanh2019distilbert,sun2019patient,wang2020minilm,jiao2020tinybert}.
For example, Sanh et al.~\shortcite{sanh2019distilbert} proposed a  DistilBERT approach, which learns the student model  from the soft target probabilities of the teacher model by using a distillation loss with softmax-temperature~\cite{jang2016categorical}, and they regularized the hidden state directions of the student and teacher models to be aligned.
Jiao et al.~\shortcite{jiao2020tinybert} proposed TinyBERT, which is an improved version of DistilBERT.
In addition to the distillation loss, they proposed to regularize the token embeddings, hidden states and attention heatmaps of both student and teacher models to be aligned via the mean squared error loss.
These methods usually learn the teacher and student models successively, where the student can only learn from the results of the teacher model.
However, the learning experience of the teacher may also be useful for the learning of student model~\cite{zhang2018deep}, which is not considered by existing methods.
In addition, the corpus for pre-training and distilling general language models (e.g., WikiPedia) may also have some domain shifts with news corpus, which may not be optimal for intelligent news applications.

In this paper, we propose a NewsBERT approach that can distill PLMs for various intelligent news applications.
In our approach, we design a teacher-student joint learning and distillation framework to collaboratively learn both teacher and student models in news intelligence tasks by sharing the parameters of top layers, and meanwhile distill the student model by regularizing the output soft probabilities and hidden representations.
In this way, the student model can learn from the teacher's  learning experience to better imitate the teacher model, and the teacher can also be aware of the learning status of the student model to enhance student teaching.
In addition, we propose a momentum distillation method by using the gradients of the teacher model to boost the gradients of student model in a momentum way, which can better transfer useful knowledge learned by the teacher model to enhance the learning of student model.
We conduct extensive experiments on two real-world datasets that involve three news intelligence tasks.
The results validate that our proposed NewsBERT approach can consistently improve the performance of these tasks using much smaller models and outperform many baseline methods for PLM distillation.

The main contributions of this work include:
\begin{itemize}
\item We propose a NewsBERT approach to distill pre-trained language models for intelligent news applications.
\item We propose a teacher-student joint learning and distillation framework to collaboratively learn both teacher and student models by sharing knowledge in their learning process.
\item We propose a momentum distillation method by using the gradient of the teacher model to boost the learning of student model in a momentum manner.
\item Extensive experiments on real-world datasets validate that our method can effectively improve the model performance in various intelligent news applications in an efficient way.
\end{itemize}

\section{Related Work}\label{sec:RelatedWork}


In recent years, many researchers explore to use knowledge distillation techniques to compress large-scale PLMs into smaller ones~\cite{tang2019distilling,sanh2019distilbert,sun2019patient,mirzadeh2020improved,sun2020mobilebert,wang2020minilm,jiao2020tinybert,wang2020minilmv2,xu2020bert,wu2021one}.
For example,  Tang et al.~\shortcite{tang2019distilling} proposed a BiLSTM$_{\rm{SOFT}}$ method that distills the  BERT model into a single layer BiLSTM using the distillation loss in downstream tasks.
Sanh et al.~\shortcite{sanh2019distilbert} proposed a DistilBERT approach, which distills the student model at the pre-training stage using the distillation loss and a cosine embedding loss that aligns the hidden states of teacher and student models.
Sun et al.~\shortcite{sun2019patient} proposed a patient knowledge distillation method for BERT compression named BERT-PKD, which distills the student model by learning from teacher's output soft probabilities and hidden states produced by intermediate layers. 
Wang et al.~\shortcite{wang2020minilm} proposed MiniLM, which employs a  deep self-attention distillation method that uses the KL-divergence loss between teacher's and student's  attention heatmaps computed by query-key inner-product and the value relations computed by  value-value inner-product.
Jiao et al.~\shortcite{jiao2020tinybert} proposed TinyBERT, which  distills the BERT model at both pre-training and fine-tuning stages by using the distillation loss and the MSE loss between the embeddings, hidden states and attention heatmaps.
There are also a few works that explore to distill pre-trained language models for specific downstream tasks such as document retrieval~\cite{lu2020twinbert,chen2020simplified}.
For example, Lu et al.~\shortcite{lu2020twinbert} proposed a TwinBERT approach for document retrieval, which employs a two-tower architecture with two separate language models to  encode the query and document, respectively.
They used the distillation loss function to compress the two BERT models into smaller ones.
These methods usually train the teacher and student models successively, i.e., distilling the student model based on a well-tuned teacher model.
However, the useful experience evoked by the teacher's learning process cannot be exploited by the student and the teacher is also not aware of the student's learning status.
In addition, the corpus for pre-training and distilling these language models usually has some domain shifts with news texts.
Thus, it may not be optimal to apply the off-the-shelf distilled language models to intelligent news applications.
In this work, we propose a \textit{NewsBERT} method to distill pre-trained language models for intelligent news applications, which can effectively reduce the computational cost of PLMs and meanwhile achieve promising performance.
We propose a teacher-student joint learning and distillation framework, where the student model can exploit the useful knowledge produced by the learning process of the teacher model.
In addition, we propose a momentum distillation method that integrates the gradient of the teacher model into the student model gradient as a momentum to boost the learning of the student.

\begin{figure*}[!t]
  \centering
    \includegraphics[width=0.65\linewidth]{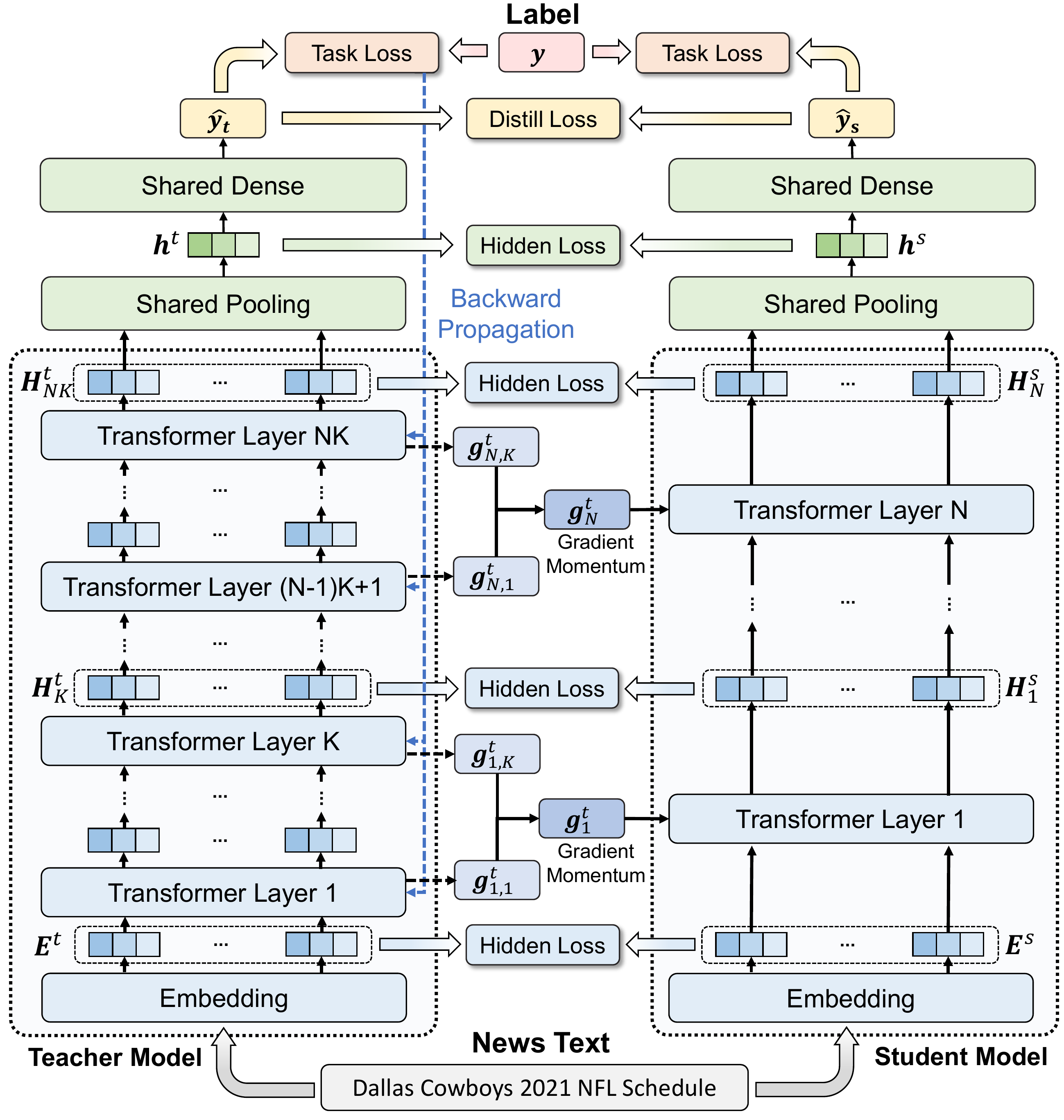}

  \caption{The framework of our \textit{NewsBERT} approach in a typical news classification task.}
  \label{fig.model}
\end{figure*}

\section{NewsBERT}\label{sec:Model}

In this section, we introduce our \textit{NewsBERT} approach that can distill PLMs for intelligent news applications.
We will first introduce the teacher-student joint learning and distillation framework of \textit{NewsBERT} by using the news classification task as a representative example, then introduce our proposed momentum distillation method, and finally introduce how to learn \textit{NewsBERT} in more complicated tasks like news recommendation.

\subsection{Teacher-Student Joint Learning and Distillation Framework}

The overall framework of our \textit{NewsBERT} approach in a typical news classification task is shown in Fig.~\ref{fig.model}.
It contains a teacher model with a parameter set $\Theta_t$ and a student model with a parameter set $\Theta_s$.
The teacher is a strong but large-scale PLM (e.g., BERT) with heavy computational cost, and the goal is to learn the light-weight student model that can keep most of the teacher's performance.
Different from existing knowledge distillation methods that first learn the teacher model and then distill the student model from the fixed teacher model, in our approach we jointly learn the teacher and student models and meanwhile distilling useful knowledge from the teacher model.
Both teacher and student language models contain an embedding layer and several Transformer~\cite{vaswani2017attention} layers.
We assume that the teacher model has $NK$ Transformer~\cite{vaswani2017attention} layers on the top of the embedding layer and the student model contains $N$ Transformer layers on the embedding layer.
Thus, the inference speed of the student model is approximately $K$ times faster than the teacher.
We first use the teacher and student models to separately process the input news text (denoted as $x$) through their Transformer layers and obtain the hidden representation of each token.
We use a shared attentive pooling~\cite{yang2016hierarchical} layer (with parameter set $\Theta_p$) to convert the  hidden representation sequences output by the teacher and student models into unified news embeddings, and finally use a shared dense layer (with parameter set $\Theta_d$) to predict the classification probability scores based on the news embedding.
By sharing the parameters of the top pooling and dense layers, the student model can get richer supervision information from the teacher, and the teacher can also be aware of student's learning status.
Thus, the teacher and student can be reciprocally learned by sharing useful knowledge encoded by them, which is helpful for learning a strong student model.

Next, we introduce the knowledge distillation details of our approach.
We assume the $i$-th Transformer layer in the student model corresponds to the layers $[(i-1)K+1, ..., iK]$ in the teacher model.
We call the stack of these $K$ layers in the teacher model as a ``block''.
Motivated by~\cite{sun2019patient}, we apply a hidden loss to align the hidden representations given by each layer in the student model and its corresponding block in the teacher model, which can help the student better learn from the teacher.
We denote the token representations output by the embedding layers in the teacher and student models as $\mathbf{E}^t$ and $\mathbf{E}^s$, respectively.
The hidden representations produced by the $i$-th layer in the student model are denoted as $\mathbf{H}^s_i$, and the hidden representations given by the corresponding block in the teacher model are denoted as $\mathbf{H}^t_{iK}$.
The hidden loss function applied to these layers is formulated as follows:
\begin{equation}\small
    \mathcal{L}^l_{hidden}(x,\Theta_t;\Theta_s)=\mathrm{MSE}(\mathbf{E}^t,\mathbf{E}^s)+\sum_{i=1}^N \mathrm{MSE}(\mathbf{H}^t_{iK},\mathbf{H}^s_i),
\end{equation}
where MSE stands for the Mean Squared Error loss function.
In addition, since the pooling layer is shared between student and teacher, we expect the unified news embeddings learned by the pooling layers in the teacher and student models (denoted as $\mathbf{h}^t$ and $\mathbf{h}^s$ respectively) to be similar.
Thus, we propose to apply an additional hidden loss to these embeddings, which is formulated as follows:
\begin{equation}\small
    \mathcal{L}^p_{hidden}(x,\Theta_t;\Theta_s,\Theta_p)=\mathrm{MSE}(\mathbf{h}^t,\mathbf{h}^s).
\end{equation}
Besides, to encourage the  student model to make similar predictions with the teacher model, we use the distillation loss function to regularize the output soft labels.
We denote the soft labels predicted by the teacher and student models as $\hat{y}_t$ and $\hat{y}_s$, respectively.
The distillation loss is formulated as:
\begin{equation}\small
    \mathcal{L}_{distill}(x,\Theta_t;\Theta_s,\Theta_p,\Theta_d)=\mathrm{CE}(\hat{y}_t/t,\hat{y}_s/t),
\end{equation}
where CE stands for the cross-entropy function and $t$ is the temperature value.
The overall loss function for distillation is a summation of the hidden losses and the distillation loss, which is formulated as:
\begin{equation}\small
    \mathcal{L}_d(x,\Theta_t;\Theta_s,\Theta_p,\Theta_d)=\mathcal{L}^l_{hidden}+\mathcal{L}^p_{hidden}+\mathcal{L}_{distill}.
\end{equation}

Since the original teacher and student models are task-agnostic, both teacher and student models need to receive task-specific supervision signals from the task labels (denoted as $y$) to tune their parameters.
Thus, the unified loss function $\mathcal{L}^s$ for training the student model is the summation of the overall distillation loss and the classification loss, which is written  as follows:
\begin{equation}\small
    \mathcal{L}^s(x,\Theta_t;\Theta_s,\Theta_p,\Theta_d)=\mathcal{L}_d(x,\Theta_t;\Theta_s,\Theta_p,\Theta_d)+\mathrm{CE}(\hat{y}_s,y).
\end{equation}
Since we do not expect the teacher to be influenced by the student too heavily, the loss function $\mathcal{L}^t$ for training the teacher model is only the classification loss, which is computed as follows:
\begin{equation}\small
    \mathcal{L}^t(x;\Theta_t,\Theta_p,\Theta_d)=\mathrm{CE}(\hat{y}_t,y).
\end{equation}
By jointly optimizing the loss functions of the teacher and student models via backward propagation, we can obtain a light-weight student model that can generate task-specific news representations for inferring the labels in downstream tasks  as the teacher model.

\subsection{Momentum Distillation}

In our approach, each Transformer layer in the student model corresponds to a block in the teacher model and we expect they have similar behaviors in learning hidden text representations.
To help the student model better imitate the teacher model, we propose a momentum distillation method that can inject the gradients of the teacher model into the student model as a gradient momentum to boost the learning of the student model.
We denote the gradients of the $j$-th layer in the $i$-th block of the teacher model as $\mathbf{g}^t_{i,j}$, which is computed by optimizing the teacher's training loss $\mathcal{L}^t$ via backward propagation.
The gradients of the $k$-th layer in the student model is denoted as $\mathbf{g}^s_{k}$, which is derived from $\mathcal{L}^s$.
We use the average of the gradients from each layer in the $i$-th block of the teacher model as the overall gradients of this block (denoted as $\mathbf{g}^t_{i}$), which is formulated as:
\begin{equation}\small
    \mathbf{g}^t_{i}=\frac{1}{K}\sum_{j=1}^K{\mathbf{g}^t_{i,j}}.
\end{equation}
Motivated by the momentum mechanism~\cite{qian1999momentum,he2020momentum},
 we combine the block gradients $\mathbf{g}^t_{i}$ with the gradients of the corresponding layer in the student model in a momentum manner, which is formulated as follows:
\begin{equation}\small
    \mathbf{g}^s_{k}=\beta \mathbf{g}^t_{k}+ (1-\beta)\mathbf{g}^s_{k}, 
\end{equation}
where $\beta$ is a momentum hyperparameter that controls the strength of the gradient momentum of the teacher model.
In this way, the teacher's gradients are explicitly injected into the student model, which may have the potential to better guide the learning of the student by pushing each layer in the student model to have similar function with the corresponding block in the teacher model.

\begin{figure}[!t]
  \centering 
      \includegraphics[width=0.98\linewidth]{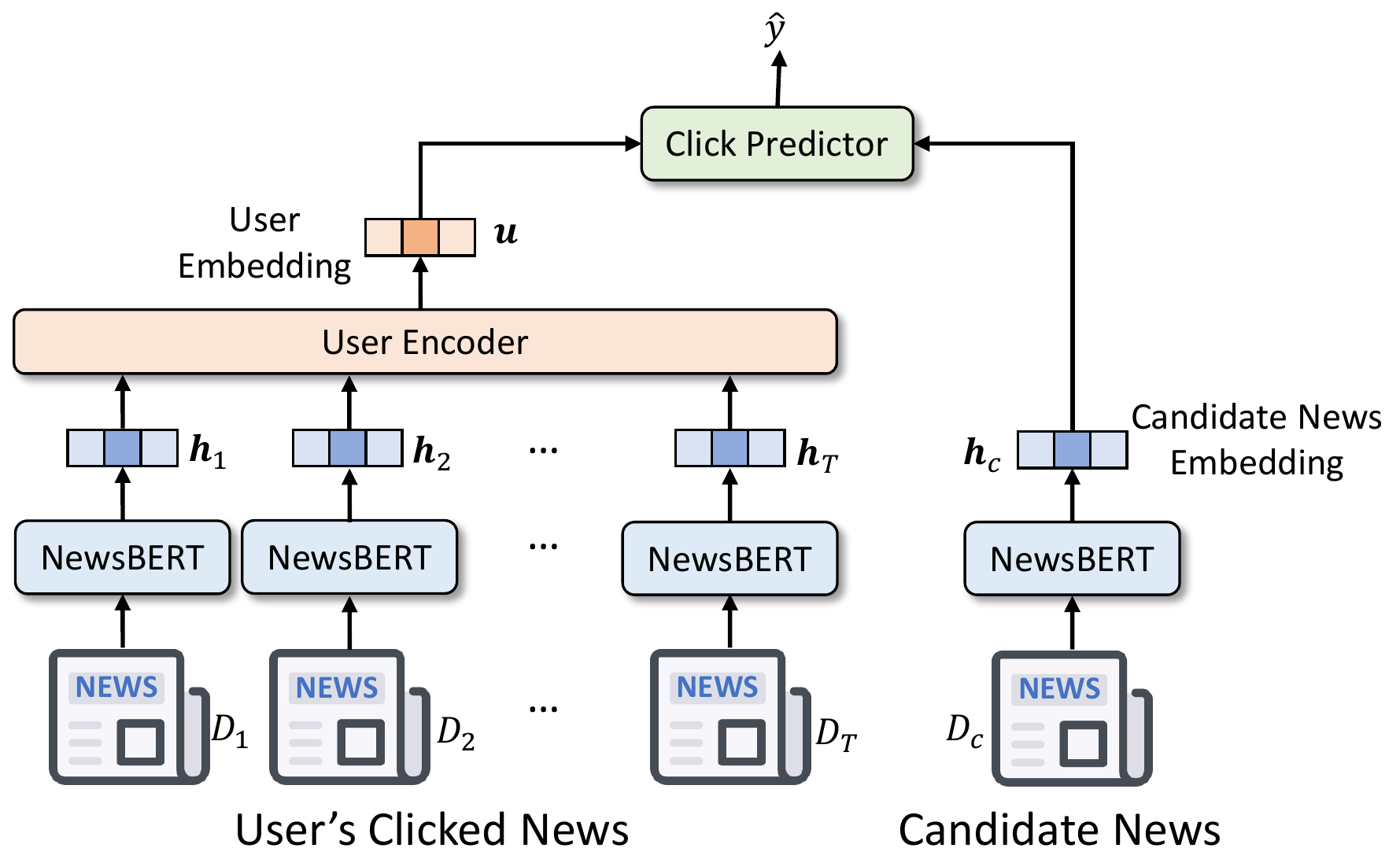}

  \caption{The framework of incorporating \textit{NewsBERT} in personalized news recommendation.}\label{fig.newsrec}

\end{figure}

\subsection{Applications of NewsBERT for News Intelligence}

In this section, we briefly introduce the applications of \textit{NewsBERT}  in other news intelligence scenarios like personalized news recommendation.
An illustrative framework of news recommendation is shown in Fig.~\ref{fig.newsrec}, which is a two-tower framework.
The input is a sequence with a user's $T$ historical clicked news (denoted as $[D_1, D_2, ..., D_T]$) and a candidate news $D_c$, and the output is the click probability score $\hat{y}$ which can be further used for personalized news ranking and display.
We first use a shared \textit{NewsBERT} model to encode each clicked news and the candidate news into their hidden representations $[\mathbf{h}_1, \mathbf{h}_2, ..., \mathbf{h}_T]$ and $\mathbf{h}_c$.
Then, we use a user encoder to capture user interest from the representations of clicked news and obtain a user embedding $\mathbf{u}$.
The final click probability score is predicted by matching the user embedding $\mathbf{u}$ and $\mathbf{h}_c$ via a click predictor, which can be implemented by the inner product function.
In this framework, teacher and student \textit{NewsBERT} models are used to generate news embeddings separately, while the user encoder and click predictor are shared between the teacher and student models to generate the prediction scores, which are further constrained by the distillation loss function.
In addition, the MSE hidden losses are simultaneously applied to all news embeddings generated by the shared \textit{NewsBERT} model and the user embedding $\mathbf{u}$ generated by the user encoder, which can encourage the student model to be similar with the teacher model in supporting user interest modeling.

\section{Experiments}\label{sec:Experiments}

\subsection{Datasets and Experimental Settings}

We conduct experiments on two real-world datasets.
The first dataset is MIND~\cite{wu2020mind}, which is a large-scale public news recommendation dataset.
It contains the news impression logs of 1 million users on the Microsoft News website during 6 weeks (from 10/12/2019 to 11/22/2019).
We used this dataset for learning and distilling our \textit{NewsBERT} model in the news topic classification and personalized news recommendation tasks.
The logs of the first 5 weeks were used for training and validation, and the rest were reserved for test.
Since many news may appear in multiple dataset splits, in the news topic classification task we only used the news that do not appear in the training and validation sets for test.
The second dataset is a news retrieval dataset (named as \textit{NewsRetrieval}), which was sampled from the logs of Bing search engine from 07/31/2020 to 09/13/2020.
It contains the search queries of users and the corresponding clicked news.
On this dataset, we finetuned models distilled on \textit{MIND} to measure their cross-task performance in news retrieval.
We used the logs in the first month for training, the logs in the next week for validation, and the rest for test.
The statistics of the two datasets are summarized in Table~\ref{dataset}.

\begin{table}[h]
\centering
\resizebox{0.48\textwidth}{!}{
\begin{tabular}{lrlr}
\Xhline{1.5pt}
\multicolumn{4}{c}{\textbf{MIND}}                                                \\ \hline
\# Users                & 1,000,000  & \# News                      & 161,013    \\
\# News categories      & 20         & \# Impressions               & 15,777,377 \\
\# Click behaviors      & 24,155,470 & Avg. \# words per news title & 11.52      \\ \hline
\multicolumn{4}{c}{\textbf{NewsRetrieval}}                                           \\ \hline
\# Queries              & 1,990,942  & \# News                      & 1,428,779  \\
Avg. \# words per query &      11.83      & Avg. \# words per news text &       596.09     \\ \Xhline{1.5pt}
\end{tabular}
}

\caption{Statistics of \textit{MIND} and \textit{NewsRetrieval}.}\label{dataset}

\end{table}

In our experiments, motivated by~\cite{chi2020infoxlm}, we used the first 8 layers of the pre-trained UniLM~\cite{bao2020unilmv2} model as the teacher model\footnote{We used the UniLM V2 model.}, and we used the parameters of its first 1, 2 or 4 layers to initialize the student models with different capacities.
In the news recommendation task, the user encoder was implemented by an attentive pooling layer, and the click predictor was implemented by inner product.
The query vectors in all attentive pooling layers were 256-dimensional.
We used Adam~\cite{kingma2014adam} as the model optimizer, and the learning rate was 3e-6.
The temperature value $t$ was set to 1.
The batch size was 32.
The dropout~\cite{srivastava2014dropout} ratio was 0.2.
The gradient momentum hyperparameter $\beta$ was set to 0.1 and 0.15 in the news topic classification task and the news recommendation task, respectively.
These hyperparamters were tuned on the validation set.
Since the topic categories in \textit{MIND} are imbalanced, we used accuracy and macro-F1 score (denoted as macro-F) as the metrics for the news topic classification task.
Following~\cite{wu2020mind}, we used the AUC, MRR, nDCG@5 and nDCG@10 scores to measure the performance of news recommendation models.
On the news retrieval task, we used AUC as the main metric.
We independently repeated each experiment 5 times and reported the average results.

\begin{table*}[!t]
\centering

\resizebox{0.7\textwidth}{!}{
\begin{tabular}{l|cc|cccc|c}
\Xhline{1.5pt}
\multicolumn{1}{c|}{\multirow{2}{*}{\textbf{Model}}} & \multicolumn{2}{c|}{\textbf{Topic classification}} & \multicolumn{4}{c|}{\textbf{News Recommendation}}                             & \multirow{2}{*}{\textbf{Speedup}} \\ \cline{2-7}
\multicolumn{1}{c|}{}                                & Accuracy                & Macro-F                 & AUC                       & MRR            & \small{nDCG@5}         & \small{nDCG@10}        &                                   \\ \hline
Glove                                               & 71.13                   & 49.71                   & 67.92                     & 33.09          & 36.03          & 41.80          & -                                 \\ \hline
BERT-12                                             & 73.68                   & 51.44                   & 69.78                     & 34.56          & 37.90          & 43.45          & 1.0x                              \\
BERT-8                                              & 73.95                   & 51.56                   & 70.04                     & 34.70          & 38.09          & 43.79          & 1.5x                              \\
UniLM-12                                            & 74.54                   & 51.75                   & 70.53                     & 35.29          & 38.61          & 44.29          & 1.0x                              \\ 
UniLM-8                                             & \textbf{74.69}             & \textbf{52.10}             & \textbf{70.72}               & \textbf{35.40}    & \textbf{38.74}    & \textbf{44.41}    & 1.5x                              \\
\hline
UniLM-4                                             & 73.53                   & 51.20                   & 69.64                     & 34.38          & 37.65          & 43.38          & 3.0x                              \\
UniLM-2                                             & 72.96                   & 50.76                   & 68.96                     & 33.52          & 36.74          & 42.48          & 6.0x                              \\
UniLM-1                                             & 72.32                   & 50.37                   & 68.02                     & 33.14          & 36.09          & 41.87          & 12.0x                    \\ \hline
TwinBERT-4*                                         & 73.59                   & 51.24                   & \multicolumn{1}{l}{69.78} & 34.48          & 37.76          & 43.47          & 3.0x                              \\
TwinBERT-2*                                         & 72.98                   & 50.84                   & 69.12                     & 33.67          & 36.89          & 42.60          & 6.0x                              \\
TwinBERT-1*                                         & 72.40                   & 50.45                   & 68.32                     & 33.40          & 36.34          & 42.09          & 12.0x                    \\ \hline
TinyBERT-6                                          & 73.54                   & 50.80                   & 69.77                     & 34.54          & 37.88          & 43.44          & 2.0x                              \\
TinyBERT-4                                          & 73.17                   & 50.39                   & 69.39                     & 33.84          & 37.50          & 43.10          & 9.4x                              \\
TinyBERT-4*                                         & 73.76                   & 51.12                   & 69.90                     & 34.52          & 37.77          & 43.48          & 3.0x                              \\
TinyBERT-2*                                         & 73.15                   & 50.94                   & 69.35                     & 33.80          & 37.42          & 43.06          & 6.0x                              \\
TinyBERT-1*                                         & 72.55                   & 50.50                   & 68.40                     & 33.46          & 36.39          & 42.15          & 12.0x                    \\ \hline
NewsBERT-4                                          & \textbf{74.45}          & \textbf{51.78}          & \textbf{70.31}            & \textbf{34.89} & \textbf{38.32} & \textbf{43.95} & 3.0x                              \\
NewsBERT-2                                          & 74.10                   & 51.26                   & 69.89                     & 34.50          & 37.75          & 43.50          & 6.0x                              \\
NewsBERT-1                                          & 73.49                   & 50.65                   & 68.97                     & 33.54          & 36.77          & 42.51          & 12.0x                    \\ 
 \Xhline{1.5pt}
\end{tabular}
}
 \caption{Performance comparisons of different methods. * Means using the UniLM model for distillation. The results of best performed teacher and student models are highlighted.} \label{table.performance} 
\end{table*}

\subsection{Performance Evaluation}
In this section, we compare the performance of our \textit{NewsBERT} approach with many baseline methods, including: (1) Glove~\cite{pennington2014glove}, which is a widely used pre-trained word embedding. We used Glove to initialize the word embeddings in a Transformer~\cite{vaswani2017attention} model for news topic classification and the NRMS~\cite{wu2019nrms} model for news recommendation.
(2) BERT~\cite{devlin2019bert}, a popular PLM with bi-directional Transformers. We compare the performance of the 12-layer BERT-Base model or its first 8 layers.
(3) UniLM~\cite{bao2020unilmv2}, a unified language model for natural language understanding and generation, which is the teacher model in our approach. We also compare its 12-layer version and its variant using the first 1, 2, 4, or 8 layers.
(4) TwinBERT~\cite{lu2020twinbert}, a method to distill PLMs for document retrieval. For fair comparison, we used the same UniLM model as our  approach.
(5) TinyBERT~\cite{jiao2020tinybert}, which is a state-of-the-art two-stage knowledge distillation method for  PLM compression. We compare the performance of the officially released 4-layer and 6-layer TinyBERT models distilled from BERT-Base and the performance of student models with 1, 2, and 4 layers distilled from the UniLM model.

Table~\ref{table.performance} shows the performance of all the compared methods  in  news topic classification and news recommendation tasks.
From the results, we have the following observations.
First, compared with the Glove baseline, the methods based on PLMs achieve better performance.
It shows that contextualized word representations generated by PLMs are more informative in language modeling.
Second, by comparing the results of BERT and UniLM (both 8- and 12-layer versions), we find UniLM-based models perform better in both tasks.
It shows that UniLM is stronger than BERT in modeling news texts, and thereby we used UniLM for learning and distilling our models.
Third, compared with BERT-12 and UniLM-12, their variants using the first 8 layers perform better.
This may be because the top layers in PLMs are adjusted to fitting the self-supervision tasks (e.g., masked token prediction) while the hidden representations of intermediate layers have better generalization ability, which is also validated by~\cite{chi2020infoxlm}.
Fourth, compared with TwinBERT, the results of TinyBERT and \textit{NewsBERT} are usually better.
This may be because the TwinBERT method only distills the teacher model based on the output soft labels, while the other two methods can also align the hidden representations learned by intermediate layers, which can help the student model better imitate the teacher model.
Fifth, our \textit{NewsBERT} approach outperforms all other compared baseline methods, and our further t-test results show the improvements are significant at $p<0.01$ (by comparing the models with the same number of layers).
This is because our approach employs a teacher-student joint learning and distillation framework where the student can learn from the learning process of the teacher, which is beneficial for the student to extract useful knowledge from the teacher model.
In addition, our approach uses a momentum distillation method that can inject the gradients of teacher model into the student model in a momentum way, which can help each layer in the student model to better imitate the corresponding part in the teacher model.
Thus, our approach can achieve better performance than other distillation methods.
Sixth, \textit{NewsBERT} can achieve satisfactory and even comparable results with the original PLM.
For example, there is only a 0.24\% accuracy gap between \textit{NewsBERT-4} and the teacher model in the topic classification task.
In addition, the size of student models is much smaller than the original 12-layer model, and their training or inference speed is much faster (e.g., about 12.0x speedup for the one-layer NewsBERT).
Thus, our approach has the potential to empower various intelligent news applications in an efficient way.

\begin{figure}[!t]
  \centering
 
      \includegraphics[width=0.9\linewidth]{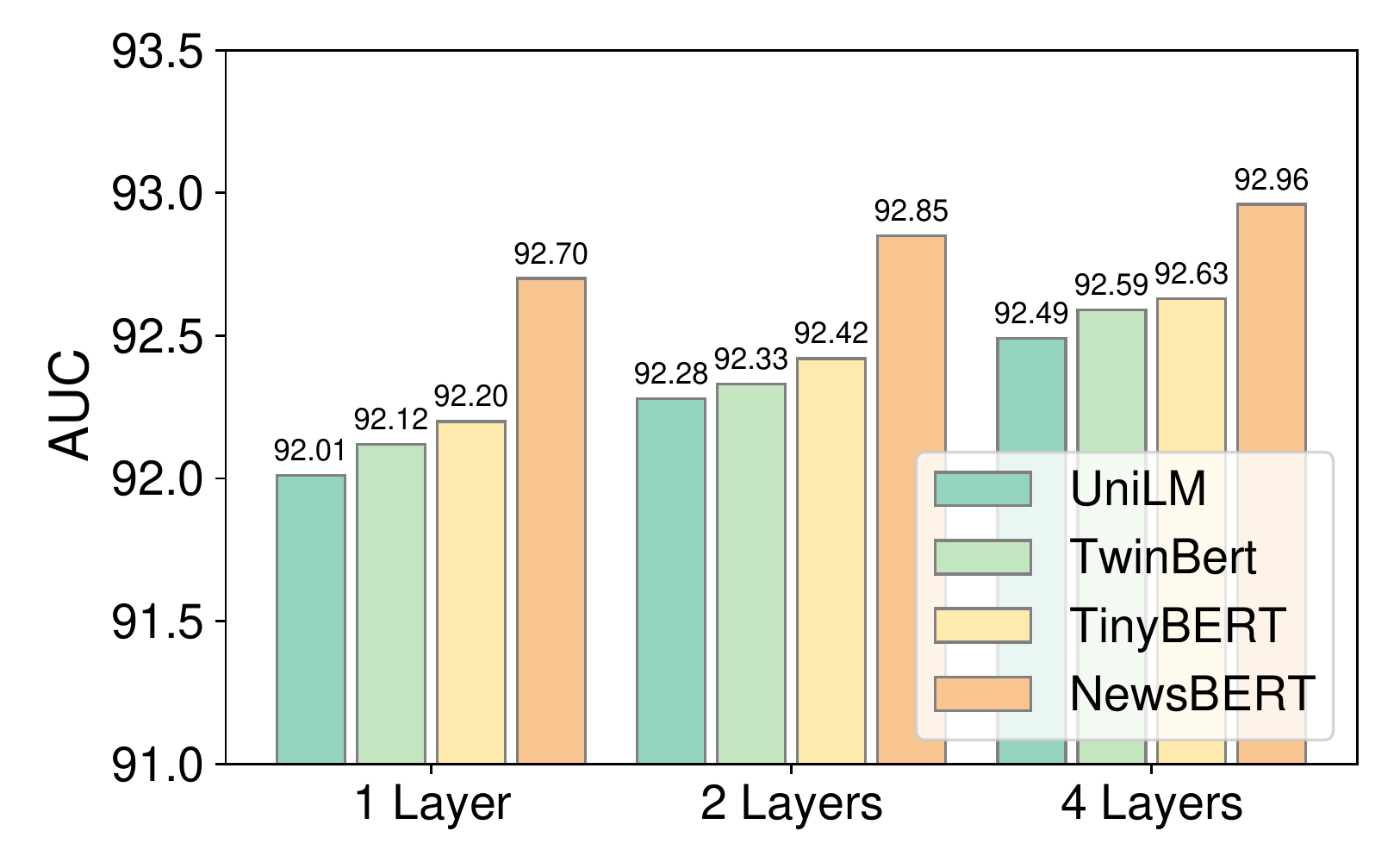}
  \caption{Cross-task performance in news retrieval.}
  \label{fig.cross}
\end{figure}

Next, to validate the generalization ability of our approach, we evaluate the performance of \textit{NewsBERT} in an additional news retrieval task.
We used the \textit{NewsBERT} model learned in the news recommendation task, and we finetuned it with the labeled news retrieval data in a two-tower framework used by TwinBERT~\cite{lu2020twinbert}.
We compared its performance with several methods, including fine-tuning the general UniLM model or the  TwinBERT and TinyBERT models distilled in the news recommendation task.
The results are shown in Fig.~\ref{fig.cross}, from which we have several findings.
First, directly fine-tuning the generally pre-trained UniLM model is worse than using the models distilled in the news recommendation task.
This is probably because that language models are usually pre-trained on general corpus like Wikipedia, which has some domain shifts with the news domain.
Thus, generally pre-trained language models may not be optimal for intelligent news applications.
Second, our \textit{NewsBERT} approach also achieves better cross-task performance than TinyBERT and TwinBERT.
It shows that our approach is more suitable in distilling PLMs for intelligent news applications than these methods.

\subsection{Effectiveness of Teacher-Student Joint Learning and Distillation Framework}

\begin{figure}[!t]
  \centering 
  \subfigure[Topic classification.]{
    \includegraphics[width=0.9\linewidth]{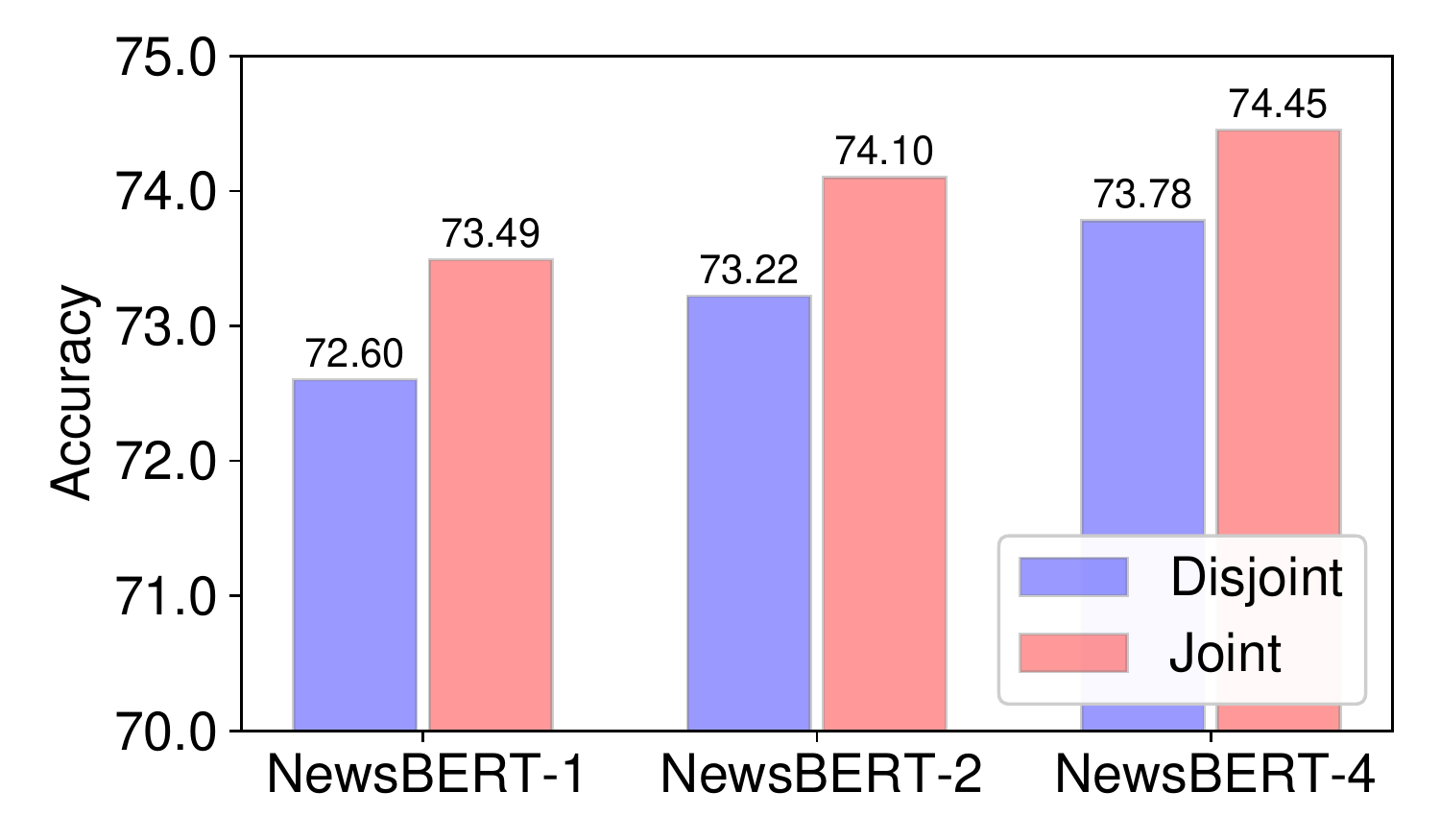} \label{fig.js1}
  }
   \subfigure[News recommendation.]{
        \includegraphics[width=0.9\linewidth]{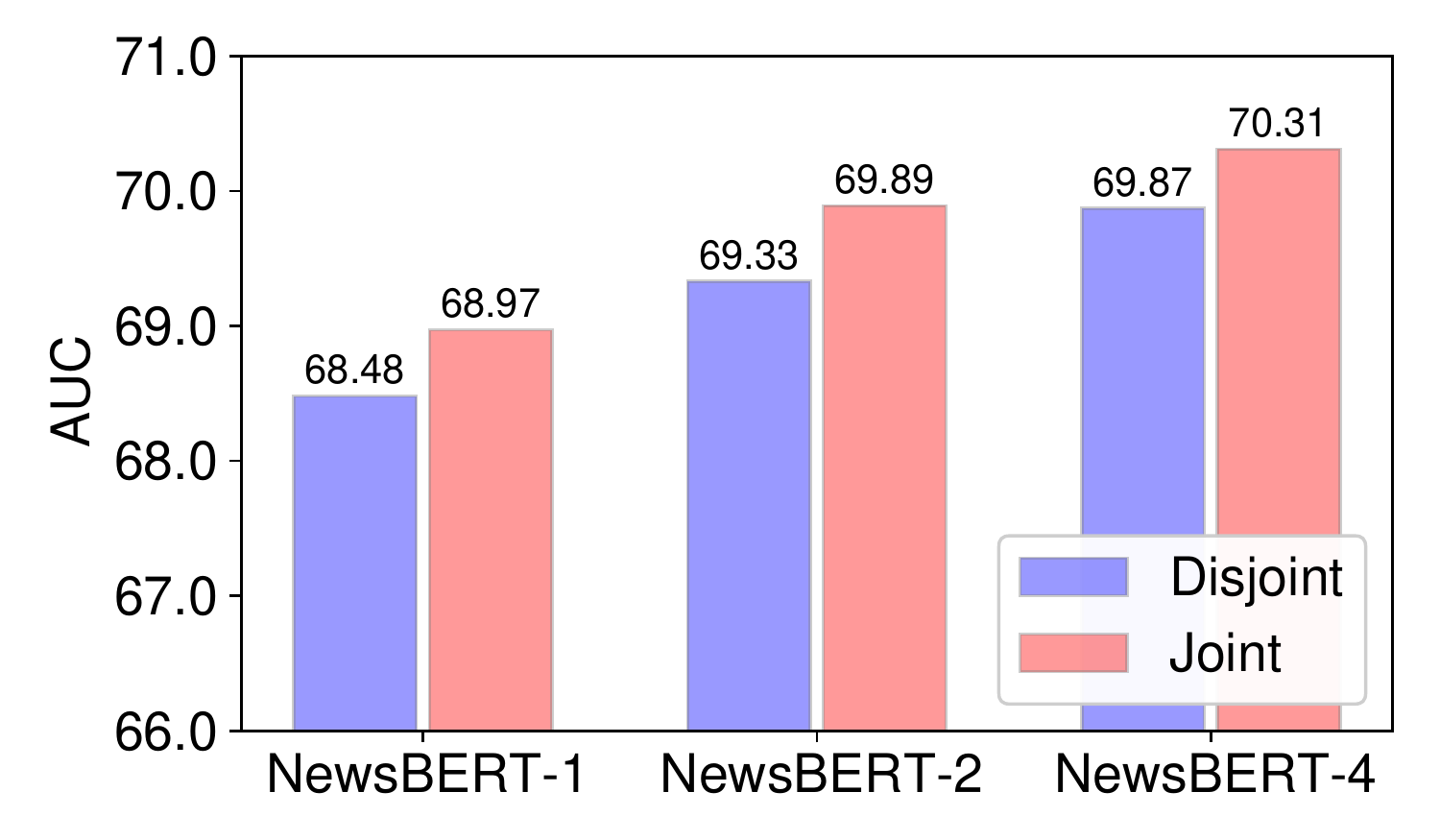} \label{fig.js2}
   }
  \caption{Influence of the teacher-student joint learning and distillation framework on the student model.}\label{fig.js}
\end{figure}

In this section, we conduct experiments to validate the advantage of our proposed teacher-student joint learning and distillation framework over  conventional methods that learn teacher and student models successively~\cite{hinton2015distilling}.
We first compare the performance of the student models under our framework and their variants learned in a disjoint manner.
The results are shown in Fig.~\ref{fig.js}.
We find that our proposed joint learning and distillation framework can consistently improve the performance of student models with different capacities.
This is because in our approach the student model can learn from the useful experience evoked by the learning process of the teacher model, and the teacher model is also aware of the student's learning status.
However, in the disjoint learning framework, student can only learn from the results of a passive teacher.
Thus, learning teacher and student models successively may not be optimal for distilling a high-quality student model.

\begin{figure}[!t]
  \centering
  \subfigure[Topic classification.]{
    \includegraphics[width=0.9\linewidth]{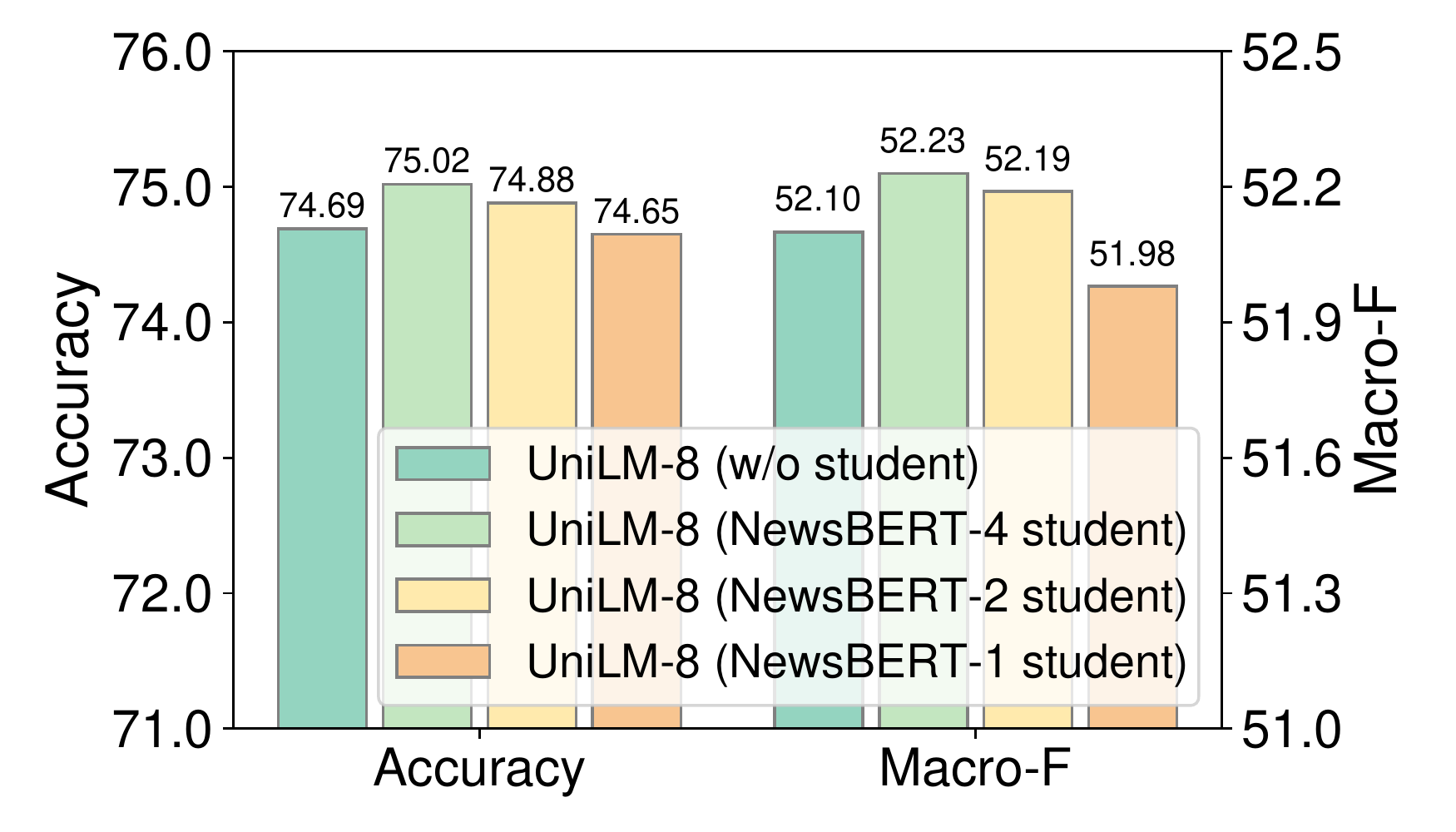}\label{fig.jt1}  
    }
    \subfigure[News recommendation.]{
      \includegraphics[width=0.9\linewidth]{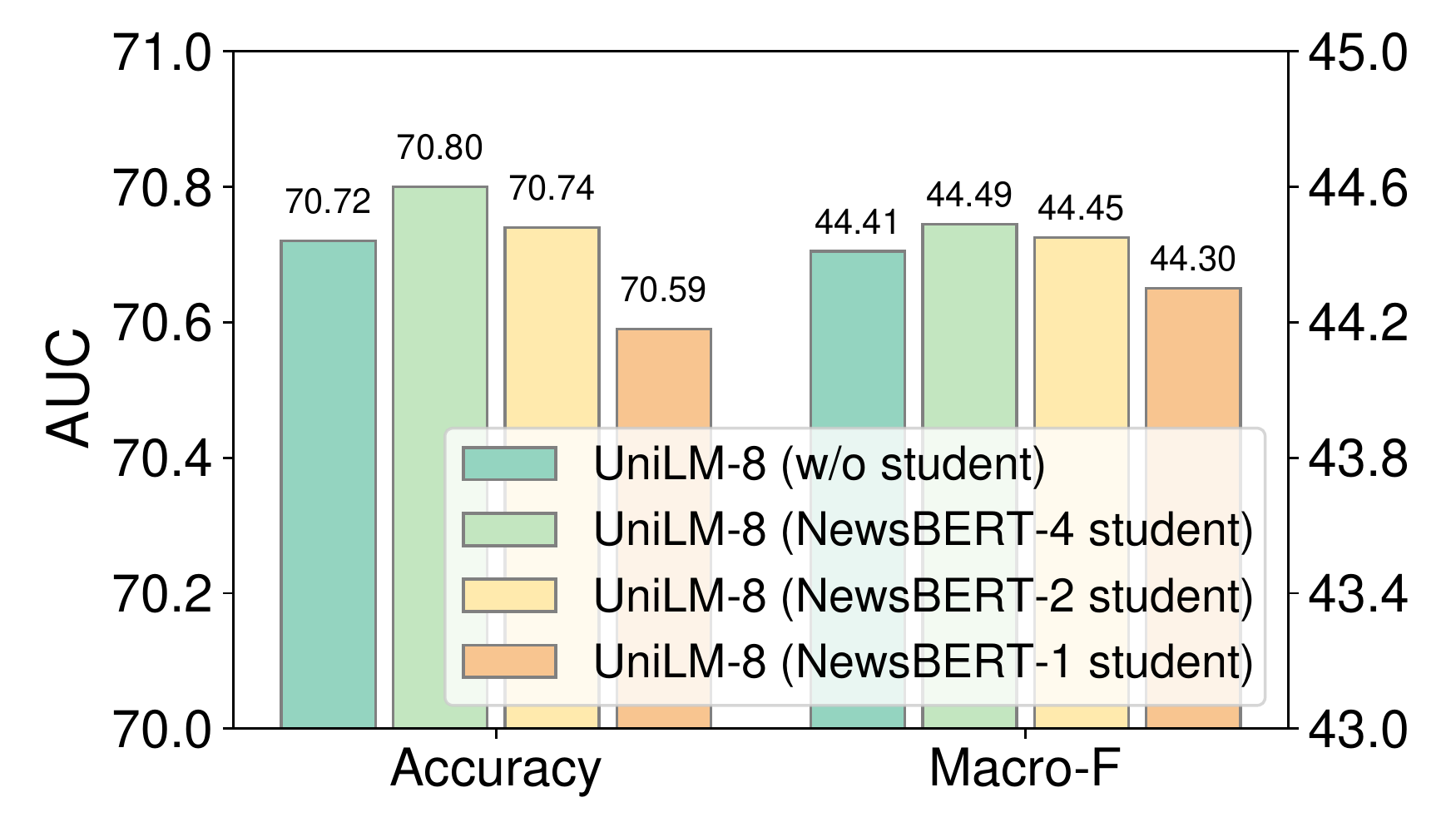}\label{fig.jt2} 
      }
  \caption{Influence of the teacher-student joint learning and distillation framework on the teacher model.}\label{fig.jt} 
\end{figure}

We also explore the influence of the teacher-student joint learning and distillation framework on the teacher model.
We compare the performance of the original UniLM-8 model and its variants that serve as the teacher model for distilling different student models.
The results are shown in Fig.~\ref{fig.jt}.
We find a very interesting phenomenon that the performance of some teacher models is better than the original  UniLM-8 model that does not participate in the joint learning and distillation framework.
This may be because the teacher model may also benefit from the useful knowledge encoded by the student model.
These results show that our teacher-student joint learning and distillation framework can help learn the teacher and student models reciprocally, which may improve both of their performance.

\subsection{Ablation Study}

\begin{figure}[!t]
  \centering 
  \subfigure[Topic classification.]{
    \includegraphics[width=0.9\linewidth]{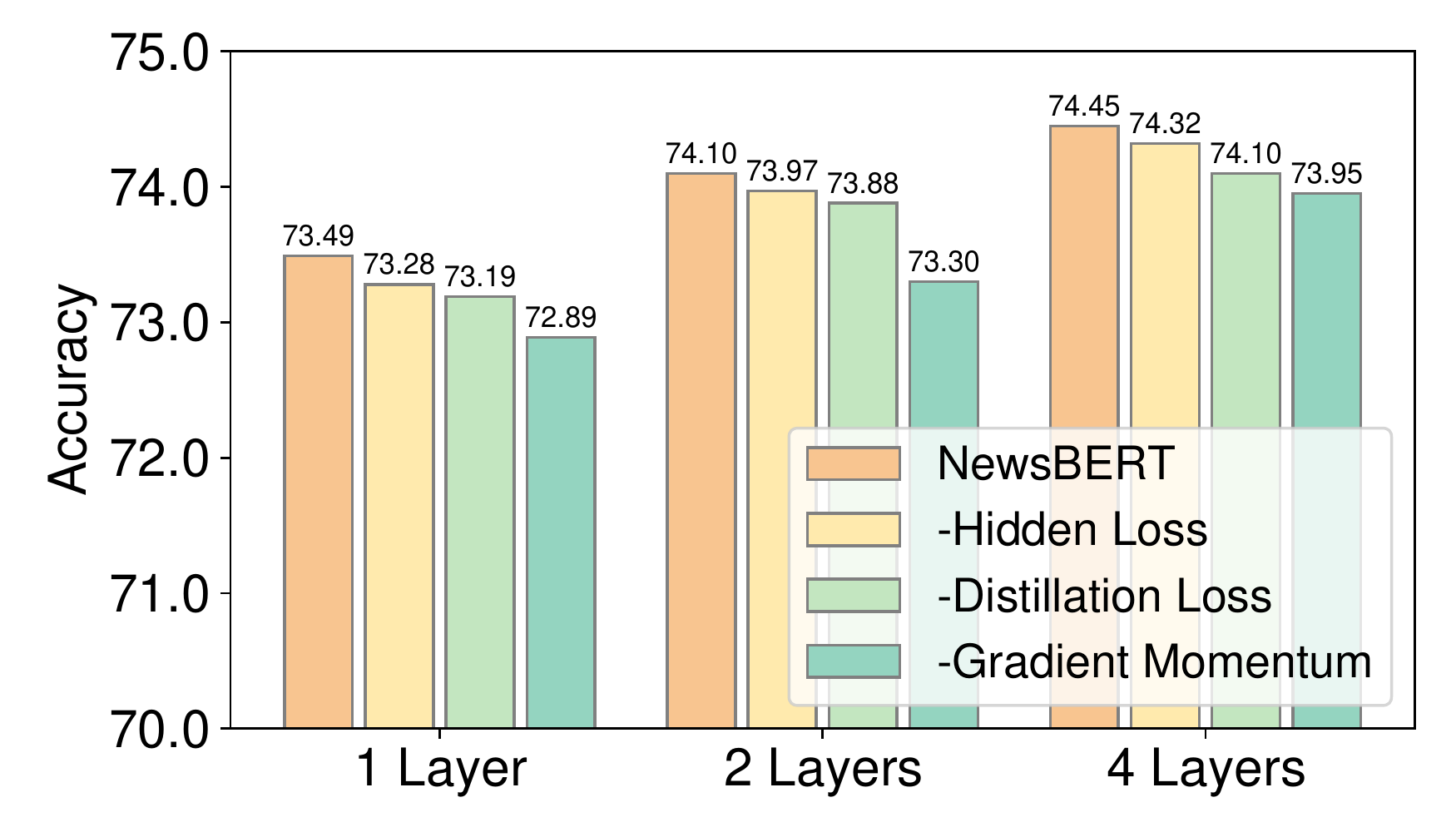} \label{fig.ab1} 
    }
    \subfigure[News recommendation.]{
       \includegraphics[width=0.9\linewidth]{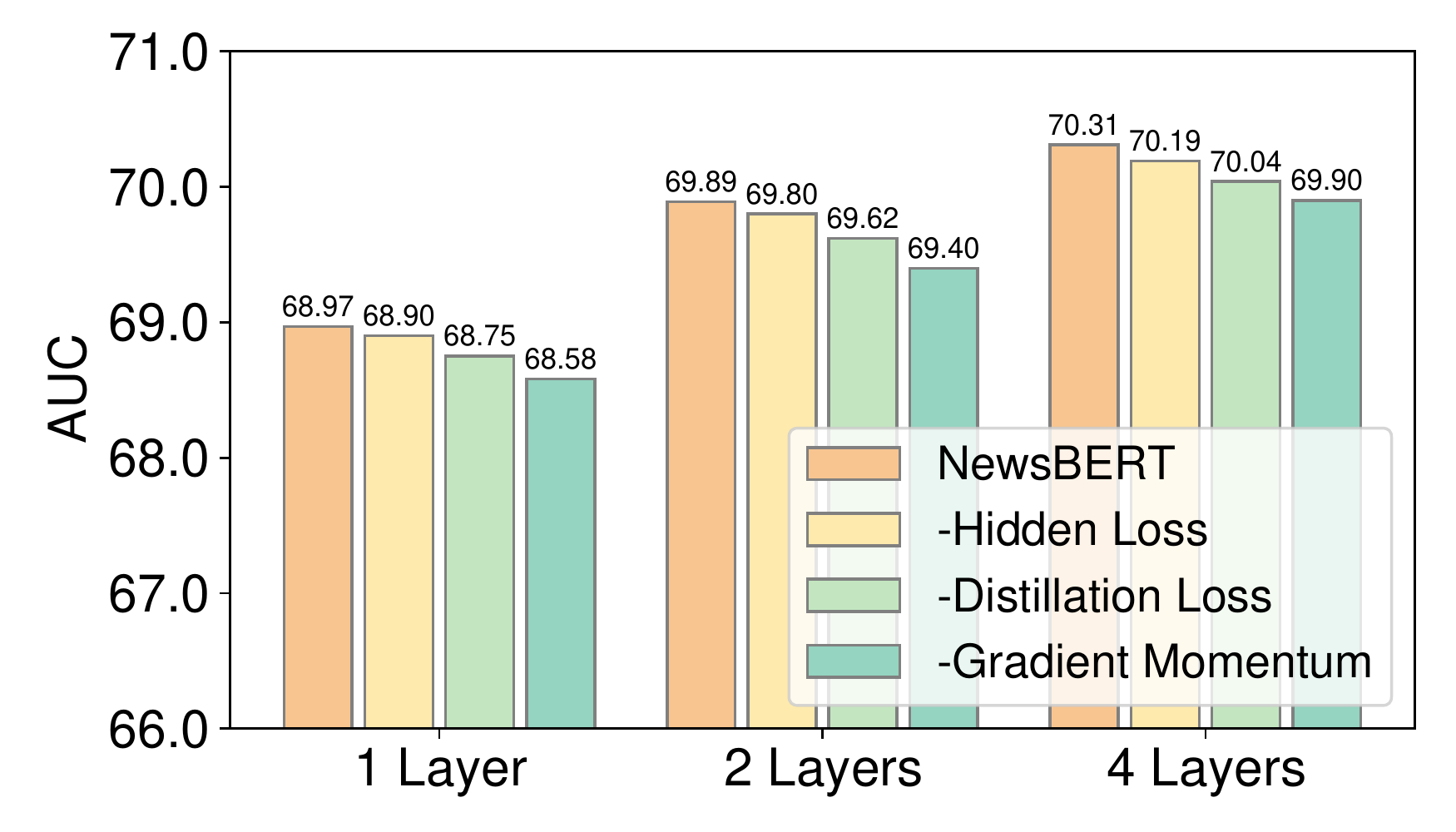}\label{fig.ab2} 
 }
  \caption{Effect of each core component in \textit{NewsBERT}.}\label{fig.ab} 
\end{figure}

In this section, we conduct experiments to validate the effectiveness of several core techniques in our approach, including the hidden loss, the distillation loss and the momentum distillation method.
We compare the performance of \textit{NewsBERT} and its variants with one of these components removed.
The results are shown in Fig.~\ref{fig.ab}.
We find that the momentum distillation method plays a critical role in our method because the performance declines considerably when it is removed.
This may be because the gradients of teacher model condense the knowledge and experience obtained from its learning process, which can better teach the student model to have similar function with the teacher model and thereby yields better performance.
In addition, the distillation loss function is also important for our approach.
This is because the distillation loss regularizes the output of the student model to be similar with the teacher model, which encourages the student model to behave similarly with the teacher model.
Besides, the hidden loss functions are also useful for our approach.
It may be because the hidden loss functions can align the hidden representations learned by the teacher and student models, which is beneficial for the student model to imitate the teacher.

\subsection{Hyperparameter Analysis}

\begin{figure}[!t]
  \centering 
  
  \subfigure[Topic classification.]{    \includegraphics[width=0.82\linewidth]{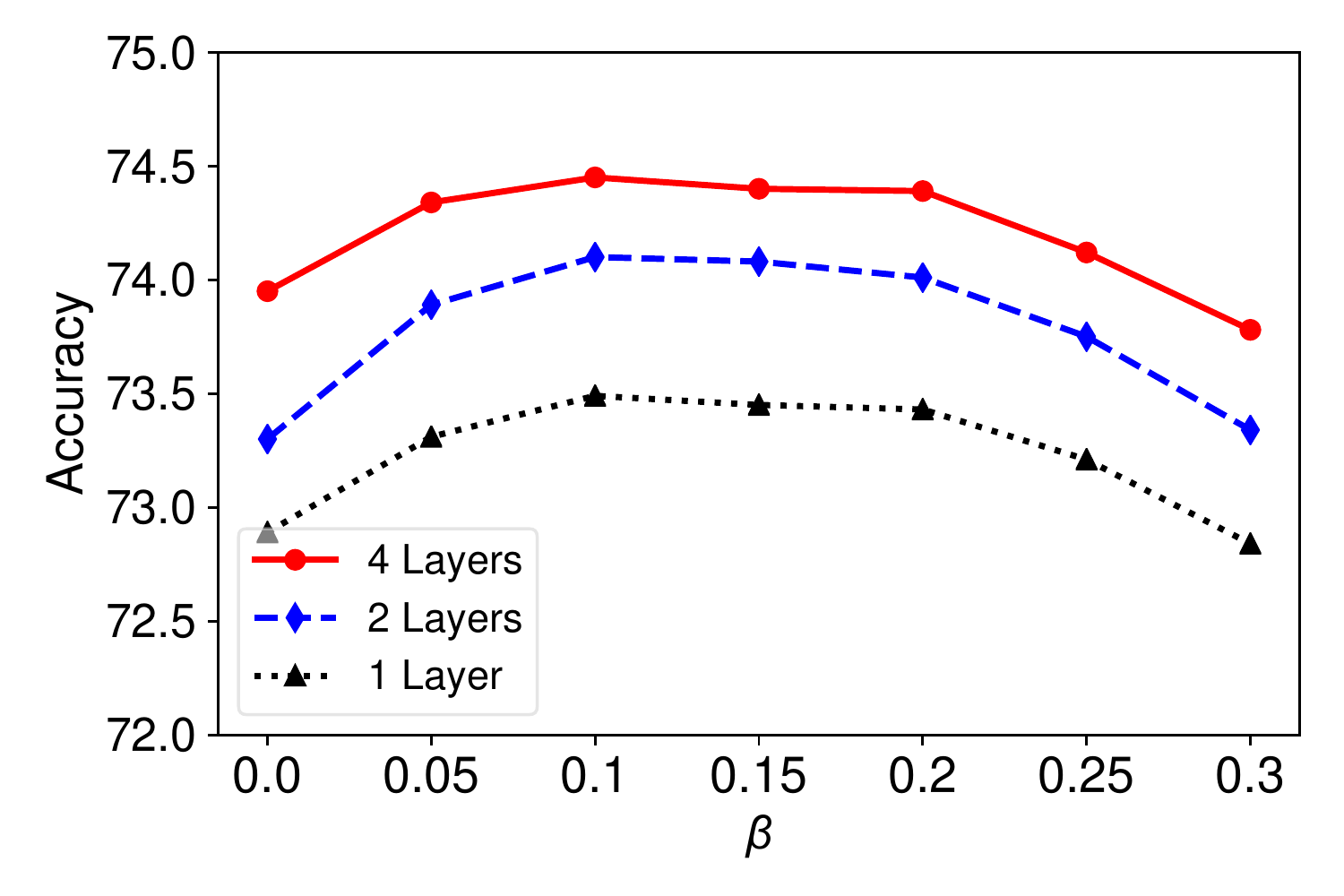}
 \label{fig.beta1}
  }
    \subfigure[News recommendation.]{ 
      \includegraphics[width=0.82\linewidth]{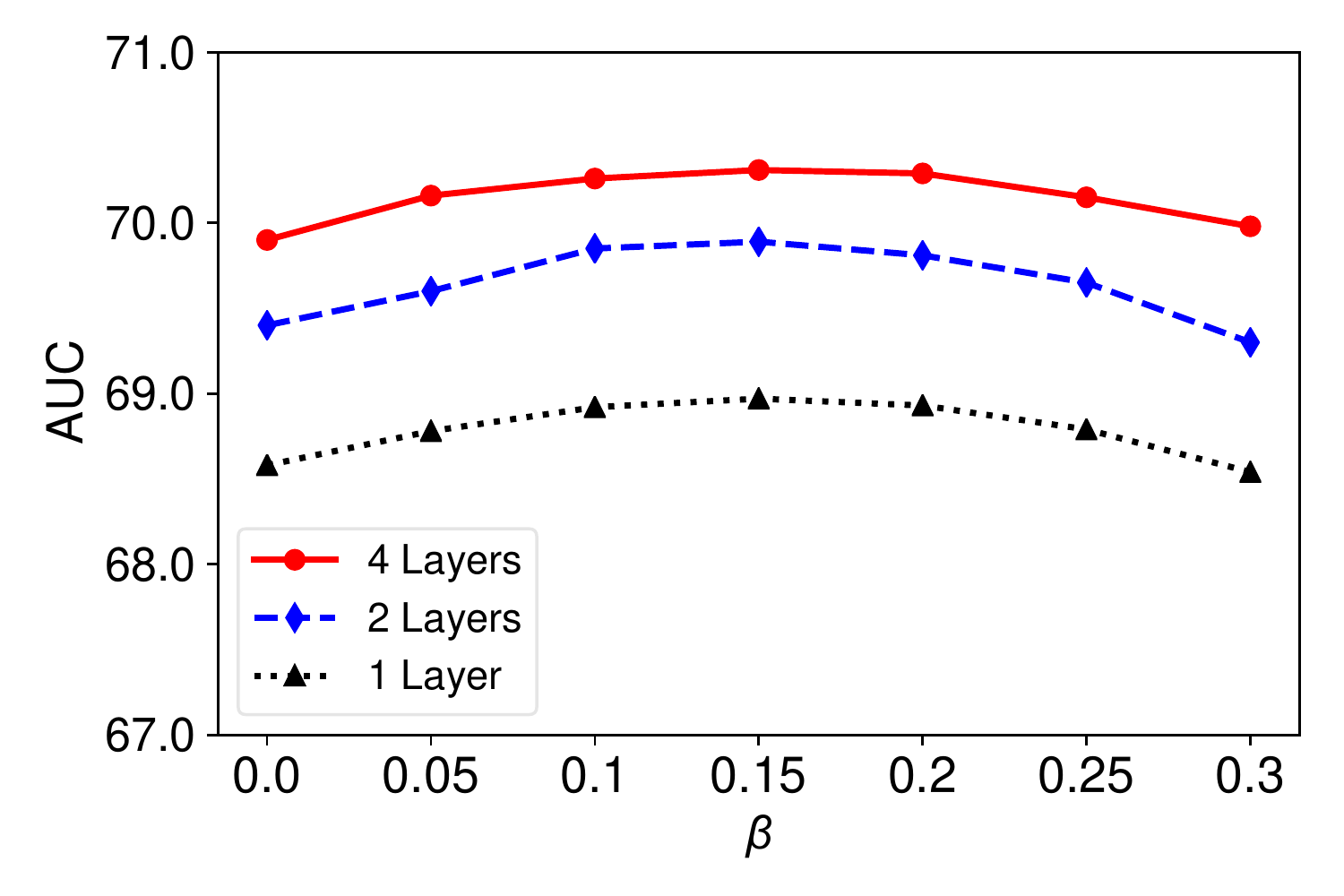}\label{fig.beta2}
  }
  \caption{Influence of the gradient momentum hyperparameter $\beta$.}\label{fig.beta}
\end{figure}

In this section, we conduct experiments to study the influence of the gradient momentum hyperparameter $\beta$ on the model performance.
We vary the value of $\beta$ from 0 to 0.3, and the results are shown in Fig.~\ref{fig.beta}. 
We observe that the performance is not optimal when the value of $\beta$ is too small.
This is because the gradient momentum is too weak under a small $\beta$, and the useful experience from the teacher model cannot be effectively exploited.
However, the performance starts to decline when $\beta$ is relatively large (e.g., $\beta>0.2$).
This is because the gradients of the teacher model inevitably have some inconsistency with the gradients of the student model, and a large gradient momentum may lead the student model updates deviate the appropriate direction.
Thus, a moderate selection of $\beta$ from 0.1 to 0.2 is recommended.

\section{Conclusion}\label{sec:Conclusion}

In this paper, we propose a knowledge distillation approach named NewsBERT to compress pre-trained language models for intelligent news applications.
We propose a teacher-student joint learning and distillation framework to collaboratively train both teacher and student models, where the student model can learn from the learning experience of the teacher model and the teacher model is aware of the learning of student model.
In addition, we propose a momentum distillation method that combines the gradients of the teacher model with the gradients of the student model in a momentum way, which can boost the learning of student model by injecting the knowledge learned by the teacher.
We conduct extensive experiments on two real-world datasets with three different news intelligence tasks.
The results show that our NewsBERT approach can effectively improve the performance of these tasks with considerably smaller models.

\section*{Acknowledgments}
This work was supported by the National Natural Science Foundation of China under Grant numbers 82090053, 61862002, and Tsinghua-Toyota Research Funds 20213930033.

\bibliography{emnlp2021}
\bibliographystyle{acl_natbib}


\end{document}